\documentclass[a4paper,12pt]{article}
\usepackage[english]{babel}
\usepackage{graphicx} 
\usepackage{framed} 
\usepackage{multirow}
\usepackage{makecell}
\usepackage{float}
\usepackage[comma]{natbib}
\setcitestyle{citesep={;}}
\usepackage{algorithm}
\usepackage{algpseudocode}
\usepackage{amsmath}
\usepackage{bm}
\usepackage{hyperref}
\usepackage[titletoc,toc,title,page]{appendix}

\begin{document}

\bibliographystyle{dcu}

\begin{center}
\begin{Large}
Analysing the Impact of Removing Infrequent Words\\[1ex] on Topic Quality in LDA Models
\end{Large}\footnote{Financial support from the German Research Foundation (DFG) (WI 2024/8-1) and
the National Science Centre (NCN) (Beethoven Classic 3: UMO-2018/31/G/HS4/00869) for the project TEXTMOD is gratefully acknowledged. The project also benefited from cooperation within HiTEC Cost Action CA 21163.}
\vspace{0.5cm}

\begin{small}
{Victor Bystrov}\\
University of Lodz\\
Rewolucji 1905r. 41, 90-214 Lodz, Poland\\
email: victor.bystrov@uni.lodz.pl\\[2ex]

{Viktoriia Naboka-Krell}\\
Justus Liebig University Giessen\\
Licher Strasse 64, 35394 Giessen, Germany\\
email: viktoriia.naboka@wirtschaft.uni-giessen.de\\[2ex]

{Anna Staszewska-Bystrova}\\
University of Lodz\\
Rewolucji 1905r. 37/39, 90-214 Lodz, Poland\\
email: anna.bystrova@uni.lodz.pl\\[2ex]

{Peter Winker}\\
Justus Liebig University Giessen\\
Licher Strasse 64, 35394 Giessen, Germany\\
email: peter.winker@wirtschaft.uni-giessen.de 
\end{small}
\end{center}

\noindent {\bf Abstract}

\noindent An initial procedure in text-as-data applications is text preprocessing. One of the typical steps, which can substantially facilitate computations, consists in removing infrequent words believed to provide limited information about the corpus. Despite popularity of vocabulary pruning, not many guidelines on how to implement it are available in the literature. The aim of the paper is to fill this gap by examining the effects of removing infrequent words for the
quality of topics estimated using Latent Dirichlet Allocation. The analysis is
based on Monte Carlo experiments taking into account different 
criteria for infrequent terms removal and various evaluation metrics. The results indicate that pruning is beneficial and that the share of vocabulary which might be eliminated can be quite considerable.\\


\newpage

\section{Introduction}

The use of topic modelling techniques, especially Latent Dirichlet Allocation (LDA) introduced by~\cite{Blei2003}, is growing fast. The methods find application in a broad variety of domains. In text-as-data applications, LDA enables the analysis of large collections of text in an unsupervised manner by uncovering latent structures behind the data.  

Given this increasing use of LDA as a standard tool for empirical analysis, also the interest in details of the method and, in particular, in parameter settings for its implementation is rising. Thus, since the introduction of the LDA approach in~\citeyear{Blei2003} by~\citeauthor{Blei2003}, different methodological components of LDA have already been studied in more detail as, for example, the choice of the number of topics~\citep{Cao2009,Mimno2011,Grossetti2022,Bystrov2022}, hyper-parameter settings~\citep{Wallach2009}, model design (e.g. hierarchical structure as proposed by~\cite{Teh2006}), and inference methods~\citep{Griffiths2004}.

However, not only the setting of technical parameters of the LDA model and the estimation algorithms are crucial for the results obtained, e.g. the identified topics. As the algorithm behind LDA ``learns'' from data provided based on co-occurrences of terms within texts, these have to be prepared in an appropriate way. LDA requires the text data to be structured in a document-term matrix (DTM), where each row corresponds to a document and each column to a specific term used throughout all documents. Then, the entry in a specific cell of the matrix provides the frequency of the term within the specific document. To obtain this matrix, the documents in a text corpus are usually cleaned and each document is represented as a bag-of-words (BoW), i.e. the algorithm neglects the semantic relationships between words and sentences. Altogether these steps are referred to as \textit{preprocessing}. By removing irrelevant terms and merging very similar terms (e.g. singular and plural forms of the same noun), preprocessing helps to reduce both the dimension of the DTM and its sparsity, which both affect the performance of the algorithms used to estimate the LDA model.

The impact of text preprocessing onto outcomes in text-as-data applications has been attracting increasing attention recently. For example, \cite{Alam_Nianmin2019} analyse the impact of different preprocessing steps on the performance of machine learning classifiers in sentiment analysis, and \cite{Barushka_Hajek2020} examine the impact of different text preprocessing settings on classifiers' performance in text classification tasks, namely recognition of fake consumer reviews. 

In contrast, in the context of LDA, no common standards seem to exist on how to perform text preprocessing. In their illustrative example, \cite{Blei2003}, for example, mention removing a standard list of stopwords and all words with an absolute frequency of one, i.e. showing up only once in the full corpus. In fact, such a step is usually performed in the majority of text-as-data applications with different lists of stopwords and alternative rules for removing low and -- sometimes also -- high frequency words. However, only few attempts have been made so far to analyse the impact of text preprocessing steps on the resulting topics. 

\cite{DennySpirling2018} address this question in their work and examine the impact of different combinations of text preprocessing steps on results obtained by unsupervised techniques including LDA (64 different specifications). Summarizing their findings, the authors highlight the importance of text preprocessing especially for unsupervised techniques such as LDA, because unlike for supervised methods, the results cannot be evaluated in a well-defined procedure (e.g. through accuracy measures as in text classification tasks). Given this limitation when using real data, the authors cannot draw more general conclusions. 

In our contribution, we focus on the impact of removing words with low frequency in the preprocessing step in the context of LDA modelling.  Usually, low frequency words make up the majority of unique terms occurring in a corpus. This feature common to many if not all languages can be approximated by Zipf's law stating in its simplest version that word frequency is proportional to the inverse of the word frequency rank. A slightly more complex model has been proposed and estimated by~\cite{Mandelbrot1953}. However, words occuring only with low frequency are believed to be too specific to contribute to the meaning of the resulting topics when applying the LDA algorithm. On the other hand, removing those words decreases the vocabulary size substantially and, consequently, accelerates model estimation. 

To the best of our knowledge, no comprehensive study has been conducted so far to analyze the impact of removing infrequent terms on topic quality. To close this research gap, we conduct a Monte Carlo (MC) simulation study. First, we define the characteristics of the data generating processes (DGPs). Following the generative model described by~\cite{Blei2003} we then create true document-topic and topic-word distributions. For each of the DGPs, we generate a total of 100 corpora using the algorithm proposed by~\cite{Bystrov2022}.\footnote{This version of the paper is preliminary regarding the limited number of replications conducted for the Monte Carlo simulations. Due to constraints in available computational resources, current work focuses on extending the number of replications to 1\,000 for the presented setup.} Finally, different techniques for defining and removing infrequent words, which have been proposed in the literature, are applied to those corpora. Afterwards, LDA models are estimated based on the preprocessed corpora. Eventually, we can analyze the impact of different settings on the model results as compared to the true DGP, document-topic and topic-word distributions.  

The remainder of this paper is structured as follows. Section~\ref{sec:preprocessing} introduces the steps usually performed for text data under the heading of text preprocessing. Focusing on removing infrequent terms, Section~\ref{sec:mc_study} describes the design of our MC study. Next, we present and discuss the results for the Monte Carlo study in Section~\ref{sec:results}. Section~\ref{sec:conclusion} concludes.  

\section{Preprocessing of Text Data}\label{sec:preprocessing}

Since texts are considered a very unstructured data source, text preprocessing usually precedes all other steps in text-as-data applications, regardless of the field of use. In general, these preprocessing steps can be divided into standard preprocessing steps and corpus or domain specific preprocessing steps. The standard steps include the following: removing punctuation, special characters, and numbers; lowercasing; removing language specific stop words; lemmatizing or stemming. This list can be adjusted or extended which can be referred to as domain specific preprocessing steps. For example, the character ``\#'' falls into the category of special characters, but keeping it can be useful when working with Twitter data. Further, the removal of extremely frequent and rare words (relative pruning) could facilitate topic modelling. 

Extremely frequent words, also called corpus-specific stop words, occur in the majority of all documents and are often considered to be insufficiently specific to be useful for topic identification. Therefore, \cite{GrimmerStewart2013} and \cite{MaierEtAl2018} remove all words that appear in more than 99\% of all documents. 

\cite{DennySpirling2018} provide two  rationals for removing very rare words: First, these words contribute little information for topics retrieval, and, second, their removal reduces the size of the vocabulary substantially and, consequently, speeds up computations. A common rule of thumb, mentioned in \cite{DennySpirling2018}, is to discard words that appear in less than 0.5-1\% of documents. \cite{DennySpirling2018} notice, however, that there has been no systematic study of effects this preprocessing choice has on topic modelling.

Infrequent terms can be removed using one of the following criteria: 
\begin{itemize} 
    \item Document frequency: remove words for which the frequency of showing up across the documents in the corpus is below the defined threshold (absolute/relative). 
    \item Term frequency: remove terms from the vocabulary the frequency of which in the corpus is below the defined threshold (absolute/relative).
    \item Term Frequency-Inverse Document Frequency (TF-IDF) values: remove words with low TF-IDF values \citep{BleiLafferty2009}.
\end{itemize}   

There are no obvious rules to set the required thresholds. \cite{GrimmerStewart2013}  notice that the choice of thresholds for removing common and rare words from a corpus should be contingent on the diversity of the vocabulary, the average length of documents and the size of the corpus. However, this is a heuristic observation that is not based on a systematic analysis.

There is almost no evidence on the impact of the removal of very common/rare words on the resulting topics. \cite{Schofield2017} address this question and conduct some experiments to test the effect of removing common words on topic quality. The experiments were conducted on two datasets, the United States State of the Union Adresses and the New York Times annotated corpus. The authors come to the conclusion that removing stop words prior to model estimation does not impact topic inference. In their experiments, they study mutual information between documents and topics to asses the effect of stopwords in topic model training.

\section{Monte Carlo Study Design}\label{sec:mc_study}

To analyse the impact of removing infrequent words in the context of LDA in a systematic way, we conduct a Monte Carlo simulation study. The goal of the MC analysis is to provide insights into the effects of vocabulary pruning on the topic quality in estimated LDA models. Given that the actual topics are known in the MC experiments, we focus, in particular, on the difference between estimated and true topics. Obviously, this difference is driven only to some extent by the specific preprocessing used, but depends also on the sampling error, which we have to take into account, when summarizing our findings.

In this section, we first describe the setup of simulation experiments. Then, we define the features that the selected DGPs should satisfy and present the procedure of corpora generation in more detail (subsection~\ref{sec:corpora}). Afterwards, we define and describe the rules for removal of  infrequent words to be tested in the MC study (subsection~\ref{sec:vocab}). Finally, we describe different quality measures used to evaluate the results (subsection~\ref{sec:evaluation}).

\subsection{Corpora Generation}\label{sec:corpora}

We start by defining two DGPs to be considered in the Monte Carlo study. Table~\ref{tab:dgps} summarizes the main characteristics of the defined DGPs. The first one contains a relatively small number of long documents covering a moderately large number of topics. These characteristics are derived from some real world datasets such as  scientific publications, reports, and speeches (e.g. \cite{HartmannSmets2018}). DGP2 covers the characteristics of corpora containing a large number of short texts discussing a  relatively small number of topics. These characteristics are typical for corpora of conference abstracts, social media, microblogs etc.   

\begin{table}[H]
\centering
\begin{tabular}{|l|c|c|c|c|}
\hline
      & \#documents & \makecell{\# words \\per document} & \makecell{\# unique terms} & \#  topics, K \\ \hline
DGP1 &    1,000    &  3,000   &   30,000   &  50  \\ \hline
DGP2 &    10,000    &  150     &   20,000    &  15  \\ \hline
\end{tabular}
\caption{DGP Characteristics}\label{tab:dgps}
\end{table}

Once we defined the characteristics of the DGP, we follow the generative model described by~\cite{Blei2003}. For each DGP the matrix of topic-word probabilities $\bm{\beta}$  is drawn from the Dirichlet distribution using a single concentration parameter $\eta=1/K$. Algorithm~\ref{alg:blei} describes how each document \textbf{w} in a corpus $D$ is generated. Document length $N$ is defined by drawing from a Poisson distribution where the parameter $\xi$ is equal to the expected number of words in a document, namely 3,000 for DGP1 and 150 for DGP2. For each document \textbf{w}, the vector of topic probabilities $\theta$ is drawn from a Dirichlet distribution using a single concentration parameter $\alpha=1/K$. For each word in a document, first a topic $z_n$ is drawn from the multinomial distribution parametrized by vector $\theta$ and then a word $w_n$ is drawn from the multinomial distribution given the topic $z_n$ and the matrix of topic-word probabilities $\bm{\beta}$. 

\begin{algorithm}[H]
    \caption{Generative probabilistic model by~\cite{Blei2003}}
Choose $\bm{\beta}\sim Dir(\eta)$ 
\begin{algorithmic}    
\For {document \textbf{w} in corpus $D$}  
    \State Choose $N \sim Poisson(\xi)$
    \State Choose $\theta \sim Dir(\alpha)$ 
    \For {word $w_{n}=1,2,\ldots,N$}
        \State (a) Choose a topic $z_{n} \sim Multinomial(\theta)$
        \State (b) {Choose a word $w_{n}$ from $p(w_{n}|z_{n}, \bm{\beta})$, a multinomial \\ \hspace{52pt} probability distribution conditioned on the topic $z_{n}$}
    \EndFor 
\EndFor
\end{algorithmic}
\label{alg:blei}
\end{algorithm}
 
Applying Algorithm~\ref{alg:blei}, we generate 100 different corpora for each DGP.

\subsection{Criteria for Removal of Infrequent Terms }\label{sec:vocab}

\subsubsection*{Document frequency}
A popular approach to  vocabulary pruning is to remove all terms that appear in a small number of documents in the corpus.
The criterion for removal can be based on the absolute (e.g., remove all terms that occur in no more than one document) or  the relative number of documents (e.g., remove all terms that occur in no more than in 1 percent of all documents in the corpus). In the MC experiments we consider different values of the relative cut-off for removing terms on the basis of relative document frequency.\footnote{We use Python's \texttt{scikit-learn} (version 0.24.2) library and its \textit{CountVectorizer()} class to approach this task.} 

\begin{figure}[H]
\centering
\includegraphics[width=\textwidth]{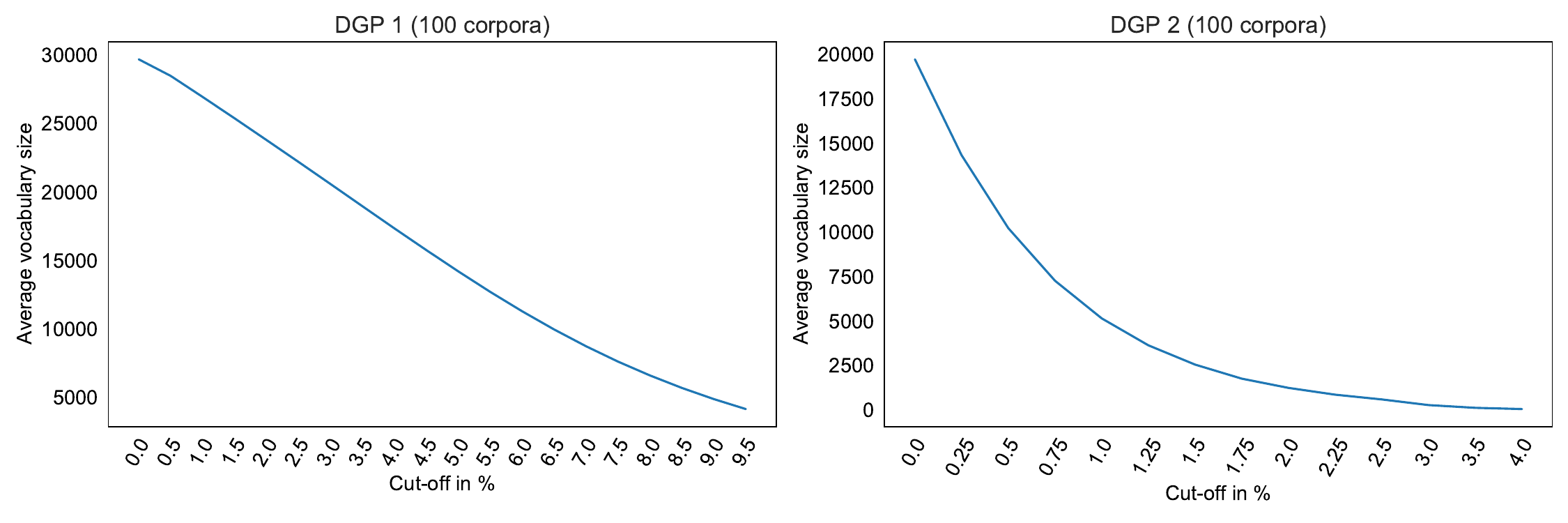}
\caption{Vocabulary size depending on the relative cut-off value}
\label{fig:vocab_size}
\end{figure}

Before fixing the range of cut-off values, we consider the resulting distribution of the vocabulary size for each DGP:  Figure~\ref{fig:vocab_size} shows the average vocabulary size as a function of the relative cut-off value (relative document based frequency). For the cut-off value of 1\% that is often used in empirical applications, the vocabulary size decreases by 9.3\% and 74.1\% on average for DGP1 and DGP2, respectively. Given these differences in the relative distributions of vocabulary sizes for selected DGPs, we use different ranges of cut-off values. For DGP1, we proceed in steps of $0.5\%$ within the interval $[0.0\%; 9.5\%]$. For DGP2, we decrease the step size to $0.25\%$ until the cut-off value of 2\% and set the maximum cut-off value to 4\% as larger cut-off values would result in an empty vocabulary.

For each of the 100 corpora based on DGP1, we build 20 different sub-samples according to the defined cut-off values and estimate one LDA model subsequently. For each of the 100 corpora from DGP2, 14 different sub-samples are constructed and corresponding LDA models are estimated.

\subsubsection*{Term frequency}
This approach to vocabulary based pruning is based on the absolute  frequency of terms in the considered corpus. We follow \cite{Blei2003} who removed all words that occurred only once in the corpus used in their illustrative example. To make the results based on term frequency comparable to the results based on document frequency, for each DGP we consider a sequence of cut-off values for absolute term frequencies such that the vocabulary size implied by each term-frequency cut-off is comparable to a vocabulary size implied by a document-frequency cut-off.  To do so, we identify the vocabulary sizes corresponding to the relative cut-offs applied in document frequency based pruning. Afterwards, we identify minimum absolute term frequencies corresponding to the considered relative cut-offs.

\subsubsection*{TF-IDF}
\cite{BleiLafferty2009} propose to use TF-IDF to prune the vocabulary. In their experiments, they consider the top 10,000 terms with highest TF-IDF values. TF-IDF is a weighted measure that is used to determine the importance of the term for the given corpus and consists of two parts, namely term frequency (TF) and inverse document frequency (IDF). 

\begin{equation}
    \text{Term Frequency}_{w,D} = \frac{\text{Number of times term $w$ appears in document $D$}}{\text{Total number of term $w$ in document $D$}}
\end{equation}

\begin{equation}
    \text{Inverse Document Frequency}_{w} = log\frac{\text{Total number of documents}}{\text{Number of documents with term $w$}}
\end{equation}

The IDF part accounts for words that occur in the majority of documents (e.g. stop words) and scales down their importance.

For each of the 100 corpora based on DGP1, we build 20 different sub-samples considering the top $V$ words with the highest TF-IDF values. To make the results comparable, we choose $V$ equal to the vocabulary size that results when document frequency-based rules are applied (see Figure~\ref{fig:vocab_size}). For example, if applying a cut-off value of 6 percent based on the relative document frequency results in a vocabulary size of about 10,000 terms, we consider only 10,000 terms with the highest TF-IDF values.

\subsection{Evaluation}\label{sec:evaluation}

Throughout each MC scenario, we keep all the parameters constant except the word-document matrix required as input for the estimation of the LDA model. As described in the previous subsection, different variations of one corpus are created by applying the defined cut-off values for removing infrequent words. As a result, for each corpus and its variations in each DGP, we obtain 20 and 14 LDA models for DGP1 and DGP2, respectively.

Different evaluation techniques have been developed to access topic modelling quality. Some of them have become standard in different text-as-data applications, e.g. topic coherence~\citep{Mimno2011} or topic similarity~\citep{Cao2009}. Perplexity is also often used to evaluate a model's predictive performance on an unseen (or held-out) sample. Perplexity is defined as the inverse of the geometric mean per-word likelihood.~\cite{Blei2003} show that perplexity is monotonically decreasing in the likelihood of the test data with increasing number of topics. Reducing the size of the vocabulary while keeping the number of topics constant leads qualitatively to the same effects. For this reason, we do not consider perplexity for evaluation in the current study.  

Instead, we further consider recall (or the share of reproduced topics) as proposed by~\cite{Bystrov2022a} and model fit to evaluate the impact of removing infrequent words on topic quality in LDA models. 

First, using the \textit{recall} metric, we aim to measure how the true structure of topics changes (\textit{true} vs \textit{estimated} topic-word distribution). In the current work, we follow a similar approach to the one proposed by~\cite{Bystrov2022a} and apply the so-called \textit{best matching}: 
\begin{enumerate}
    \item Combine true and estimated word-topic distributions based on the union of the two vocabularies. For words not contained in the estimated word-topic distribution, assign the probability of zero. In doing so, we obtain vectors of the same length. An example of this procedure is presented in Figure~\ref{fig:best_matching_example} below. 
    \begin{figure}[H]
        \centering
        \includegraphics[width = 0.8\textwidth]{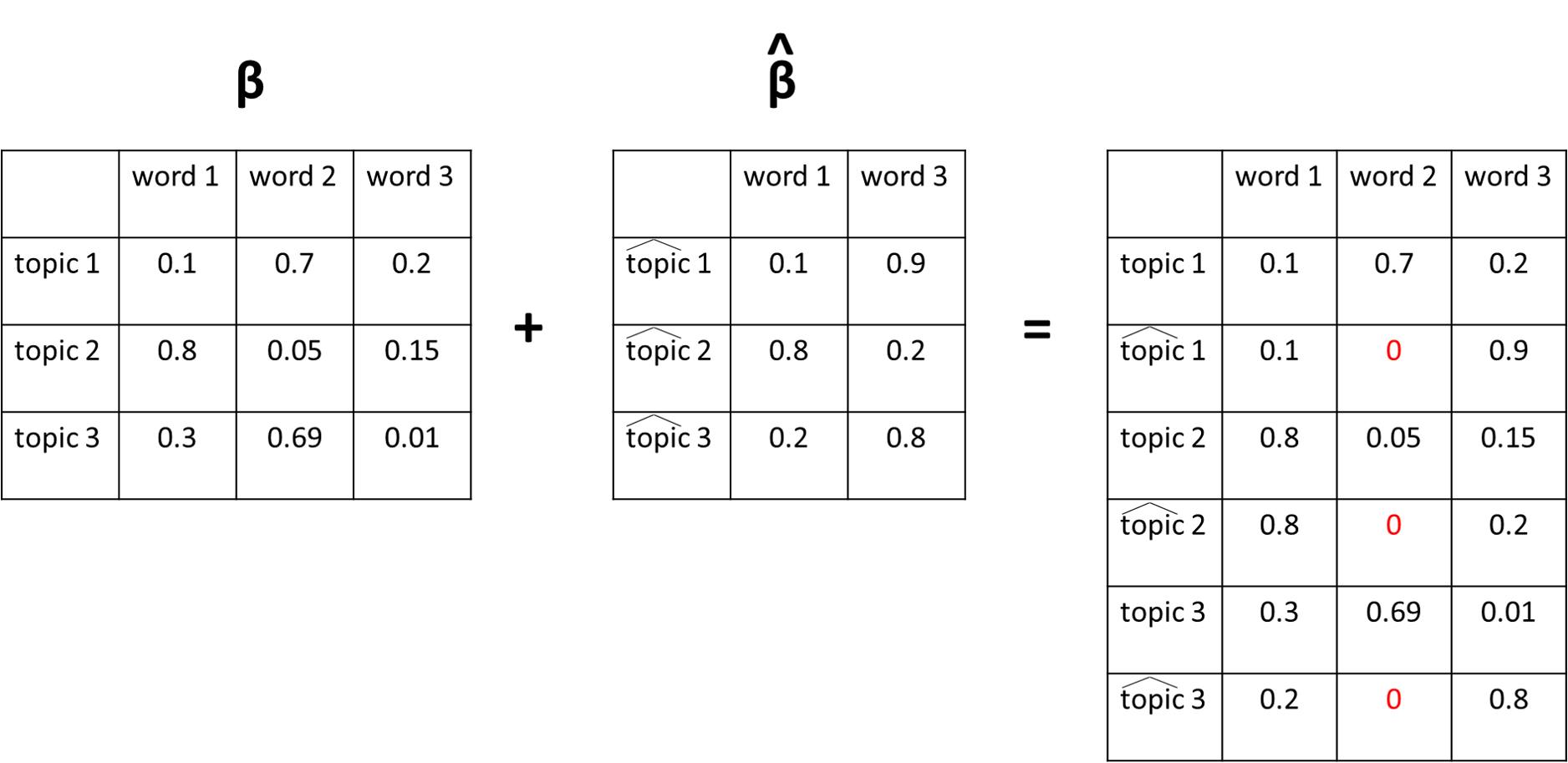}
        \caption{Best Matching: example}
        \label{fig:best_matching_example}
    \end{figure}
    \item For each of the estimated topics, calculate \textit{similarity/distance} to each of the true topics. Then, assign the topic with the highest (lowest) similarity (distance). 
    \item Define and apply a cut-off value to keep only sensible matches. The \textit{recall} metric is then calculated as the share of correctly reproduced topics.  
\end{enumerate}

In their empirical application, \cite{Bystrov2022a} use cosine similarity in step 2 and automatically determine a data based cut-off as the 95\% percentile of all pairwise similarity scores in step 3. \cite{MaierEtAl2020}, who also studied the impact of removing infrequent terms on topic quality, perform topic matching based on top 20 topic words following the approach proposed by~\cite{Niekler2012}. The authors calculate pairwise cosine distances and apply a cut-off value of 0.5 to obtain the share of reproduced topics.

In the current application, we use different metrics to measure the similarity between true and estimated topics:
    \begin{itemize}
        \item Cosine similarity: ranges between 0 (two vectors orthogonal) and 1 (vectors are pointing in the same direction).
        \item Jensen-Shannon divergence/distance: ranges between 0 (two distributions are the same) and 1 (completely different).
        \item Rank-Biased Overlap (RBO) proposed by~\cite{Webber} is a similarity metric to compare ranked lists. It ranges from 0 (disjoint) to 1 (exactly the same).
    \end{itemize}
Since the true topics appear to be very distinct from each other in the current MC study, we decide to use an ad-hoc cut-off value of 0.8 for the similarity metrics (cosine similarity and RBO) and 0.2 for the distance metric (Jensen-Shannon).

Alternatively, one can use \textit{one-to-one matching} also proposed by~\cite{Bystrov2022a}. Thereby, all of the topics have to be matched using the Hungarian algorithm and a defined distance metric. Matches are assigned to minimize the overall cost of assignment. Thus, the mean of the distances between the identified matches can be considered to measure the quality of model fit.

\section{Results}\label{sec:results}

\begin{figure}[H]
\centering
\includegraphics[width=0.85\textwidth]{"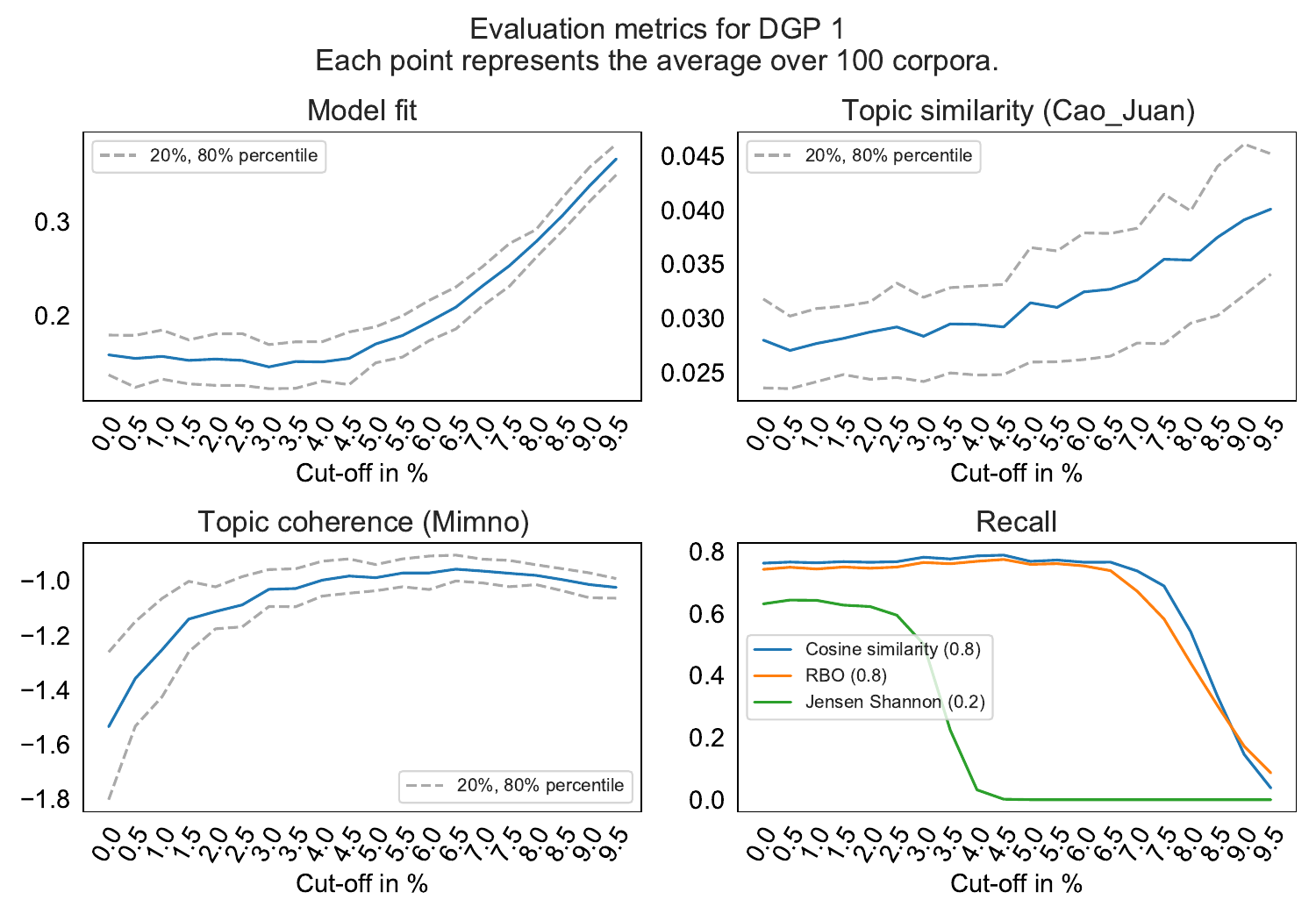"}
\caption{Evaluation of document frequency-based vocabulary pruning for DGP1}
\label{fig:dgp1_metrics1}
\end{figure}

\begin{figure}[H]
\centering
\includegraphics[width=0.85\textwidth]{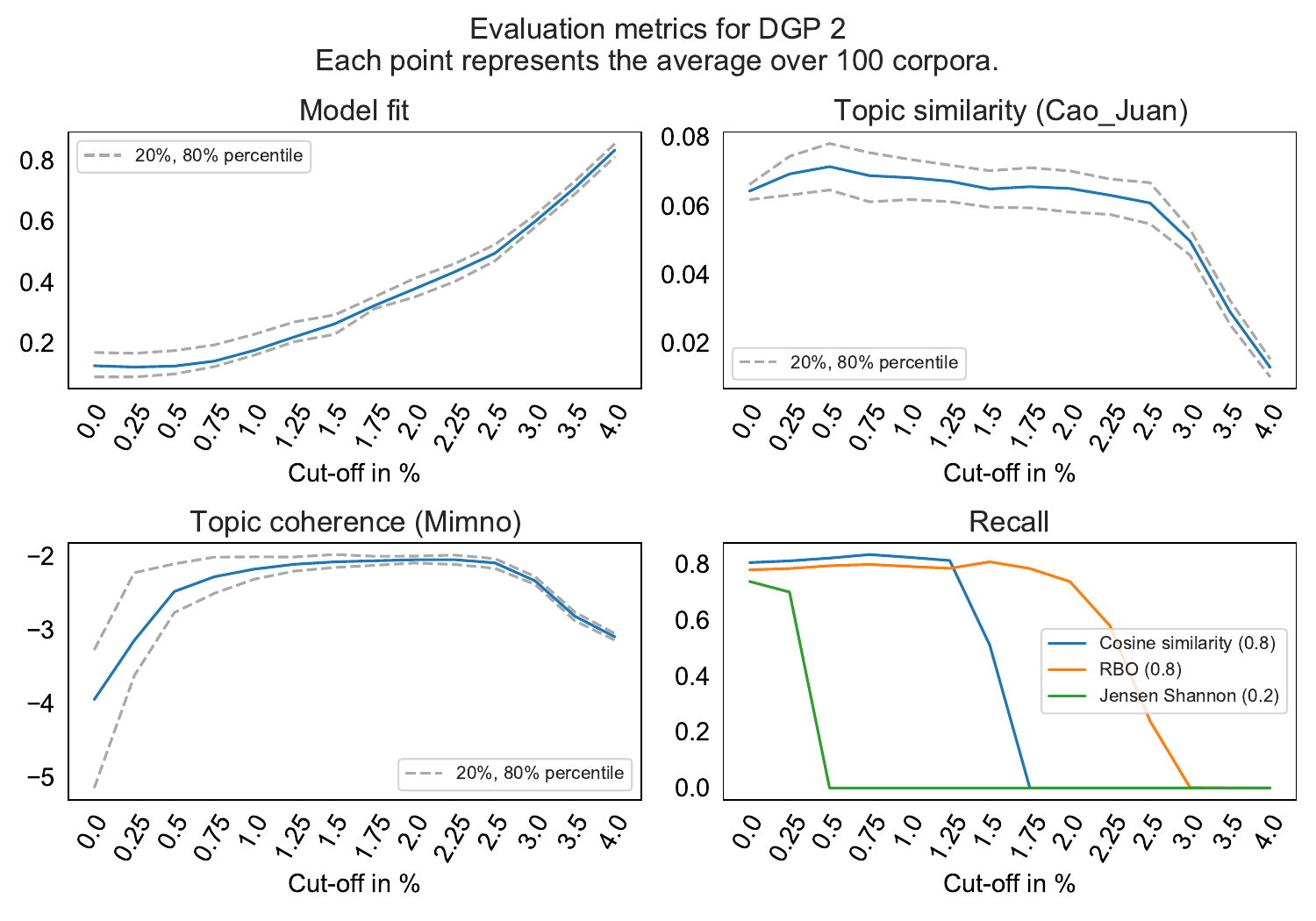}
\caption{Evaluation of document frequency-based vocabulary pruning for DGP2}
\label{fig:dgp2_metrics1}
\end{figure}

In this section, we summarize the main findings of the Monte Carlo analysis. Thereby, we focus on the removal of infrequent terms according to their document frequency in the corpus.\footnote{Results obtained for the alternative criteria for vocabulary pruning, described in Section~\ref{sec:vocab}, are presented in Appendix~\ref{app:reducing_vocab}.} The cut-off values exhibited on the $x$-axis in Figures~\ref{fig:dgp1_metrics1} and~\ref{fig:dgp2_metrics1} describe the minimum share of documents a term has to be included in for not being removed from the corpus. Thus, a cut-off value of 0.0\% corresponds to keeping all terms (30K for DGP1 and almost 20K for DGP2), while 9.5\% in Figure~\ref{fig:dgp1_metrics1} refers to the removal of all terms which do not show up in at least 9.5\% of all documents leaving only about 4K terms in the corpus. Accordingly, in Figure~\ref{fig:dgp2_metrics1}, the value of 4.0\% corresponds to keeping only those terms, which show up at least in 4.0\% of all documents reducing the size of the vocabulary to 60 terms. 

On the ordinate, Figures~\ref{fig:dgp1_metrics1} and~\ref{fig:dgp2_metrics1} show the means of the evaluation metrics obtained over 100 replications for DGP1 and DGP2, respectively, as solid lines. 
The dashed lines in the first three subplots provide the 20\% and 80\% quantiles of these distributions. Corresponding bands for measures from the last panel (\textit{recall}) are shown in Figures~\ref{fig:dgp1_recall_bands} and~\ref{fig:dgp2_recall_bands} from Appendix~\ref{appendixb}.
The metrics under consideration include: \textit{model fit}~\citep{Bystrov2022a} (to be minimized), topic \textit{similarity}~\citep{Cao2009} (to be minimized), topic \textit{coherence}~\citep{Mimno2011} (to be maximized), and \textit{recall} (to be maximized). In empirical applications, the true DGPs and corresponding topics are unknown. Thus, the \textit{recall} criteria cannot be applied. The observed collapse of \textit{recall} for higher cut-off values indicates that the remaining vocabulary is not sufficient anymore for identifying the true topics. 

It becomes obvious from Figures~\ref{fig:dgp1_metrics1} and~\ref{fig:dgp2_metrics1} that removing infrequent terms has consequences for LDA estimation results. As a general pattern we conclude that applying pruning is always beneficial for low cut-off values. This might be attributed to two effects. First, terms showing up only in a few documents do not contain much information about more general topics. Second, removing these terms reduces the dimensionality of the estimation problem substantially, which increases the efficiency of the estimators. However, beyond a certain point the increasing loss of information resulting from the removal of more and more infrequently used terms dominates the gains due to reduced dimensionality. Comparing the findings from Figures~\ref{fig:dgp1_metrics1} and~\ref{fig:dgp2_metrics1}, it appears that gains and losses from decreasing vocabulary size by eliminating rare terms are weighted somewhat differently by alternative evaluation criteria.  

For DGP1 (Figure~\ref{fig:dgp1_metrics1}), the lowest average distances between the true and estimated topic sets as measured by \textit{model fit} correspond to cut-off values from 3\% to 4.5\%. Further removal of infrequent terms leads to increasing distortions in estimated topics. The best values of \textit{coherence} are obtained for thresholds of 3\%-6.5\%. The metric is quite sensitive to keeping too many infrequent terms in the texts, showing significantly smaller values for initial thresholds. The best cut-off value indicated by topic \textit{similarity} is 0.5\%. Nevertheless, thresholds up to 4.5\% lead to similar values of the metric. Eventually, alternative versions of \textit{recall} measures indicate that the maximum threshold which might be considered is about 3\% (metric based on the Jensen-Shannon distance) or 6.5\% (cosine similarity and RBO based metrics). Altogether, if all metrics are considered jointly, the best threshold is about 3\%. A similar conclusion is reached if TF-IDF or absolute term frequency based vocabulary pruning is performed instead of document frequency pruning (see Appendix~\ref{app:reducing_vocab}).

A similar analysis for DGP2 (Figure~\ref{fig:dgp2_metrics1}) suggests the following cut-off values. According to \textit{model fit} the interval from 0.25\% to 0.75\% might be considered, while the \textit{coherence} metric indicates the range 0.5\%-2.5\%. Topic \textit{similarity} is quite similar for cut-off values up to 1.25\% and \textit{recall} metrics suggest to stop at 0.25\%, 1.25\% or 2\% starting from the most restrictive measure. Thus, overall a threshold of about 0.25\%-0.5\% might be selected. This finding is again quite robust with respect to the definition used for infrequent terms removal (see Appendix~\ref{app:reducing_vocab}).

For a better understanding of the results from Figures~\ref{fig:dgp1_metrics1} and \ref{fig:dgp2_metrics1}, the selected thresholds where juxtaposed with the corresponding shares of words removed from the vocabularies (see Figures~\ref{fig:dgp1_recall_add_axes} and~\ref{fig:dgp2_recall_add_axes} in Appendix~\ref{appendixb}). The cut-off value of 3\% for DGP1 corresponds to shrinking the size of the vocabulary by 30\% and cut-offs of 0.25-0.5\% for DGP2 imply removing 27-48\% of all terms. Thus, in both cases it could be concluded that the reduction of the size of the vocabulary, which could accelerate the estimation process without affecting the results qualitatively, was considerable and amounted to about 30\% of all terms. These findings show that guidelines focusing on removing infrequent terms up to a certain share of all terms might be worth following up.

\section{Conclusion}\label{sec:conclusion}
The focus of this paper was on preprocessing of text data in the context of LDA model estimation. Although text preprocessing is an essential part of data preparation in text-as-data applications and some rules-of-thumb of text preprocessing sequences exist and are often followed, there is only little evidence on how particular text preprocessing decisions might affect the final results. In the specific setting considered in this paper, the outcome of interest were the resulting estimated topics and the analysed preprocessing step was the removal of infrequent words in a text corpus. 

To allow for a systematic evaluation of the impact of different techniques on reducing vocabulary size and generalizable conclusions, we conducted a Monte Carlo simulation study. We first generated data from scratch based on two pre-defined DGPs following the probabilistic model proposed by~\cite{Blei2003}. For each of the defined DGPs, we then applied different techniques to remove rare words from the texts and estimated several LDA models varying the text input only. Finally, we evaluated results using some well established metrics such as \textit{coherence} and \textit{topic similarity} that focus on the estimated set of topics as well as \textit{model fit} and \textit{recall} metrics that focus on the comparison between true and estimated set of topics.   

The results of the current paper have at least two practical implications. 
First, it has been shown that across the considered DGPs about 30\% of words can be removed without qualitative losses in the resulting topics. This is a valuable insight for the scientists who work with substantial amount of data containing long texts on average. Most real-world data sets have large or even very large vocabularies. In such cases, removing 30\% of words could result in a considerable decrease in computing time and an increase in efficiency. Second, we performed robustness checks applying different techniques to reduce the size of vocabularies, e.g., TF-IDF, absolute frequency. The outcomes of different techniques were made comparable by controlling for the resulting size of the remaining vocabulary. Independent of the applied procedure, we come to similar conclusions.    

Based on the results of the current study, future research could follow the ideas of~\cite{DennySpirling2018} and focus on different combinations of text preprocessing steps and investigate their impact in a systematic manner by conducting further Monte Carlo studies, which would require, however, substantially more computational resources. For example, it might be worthwhile to consider the combined impact of stemming/lemmatizing and vocabulary pruning.

\bibliography{literature}

\begin{appendices}
\section{Robustness Checks}\label{app:reducing_vocab}

\begin{figure}[H]
\centering
\includegraphics[width=0.85\textwidth]{"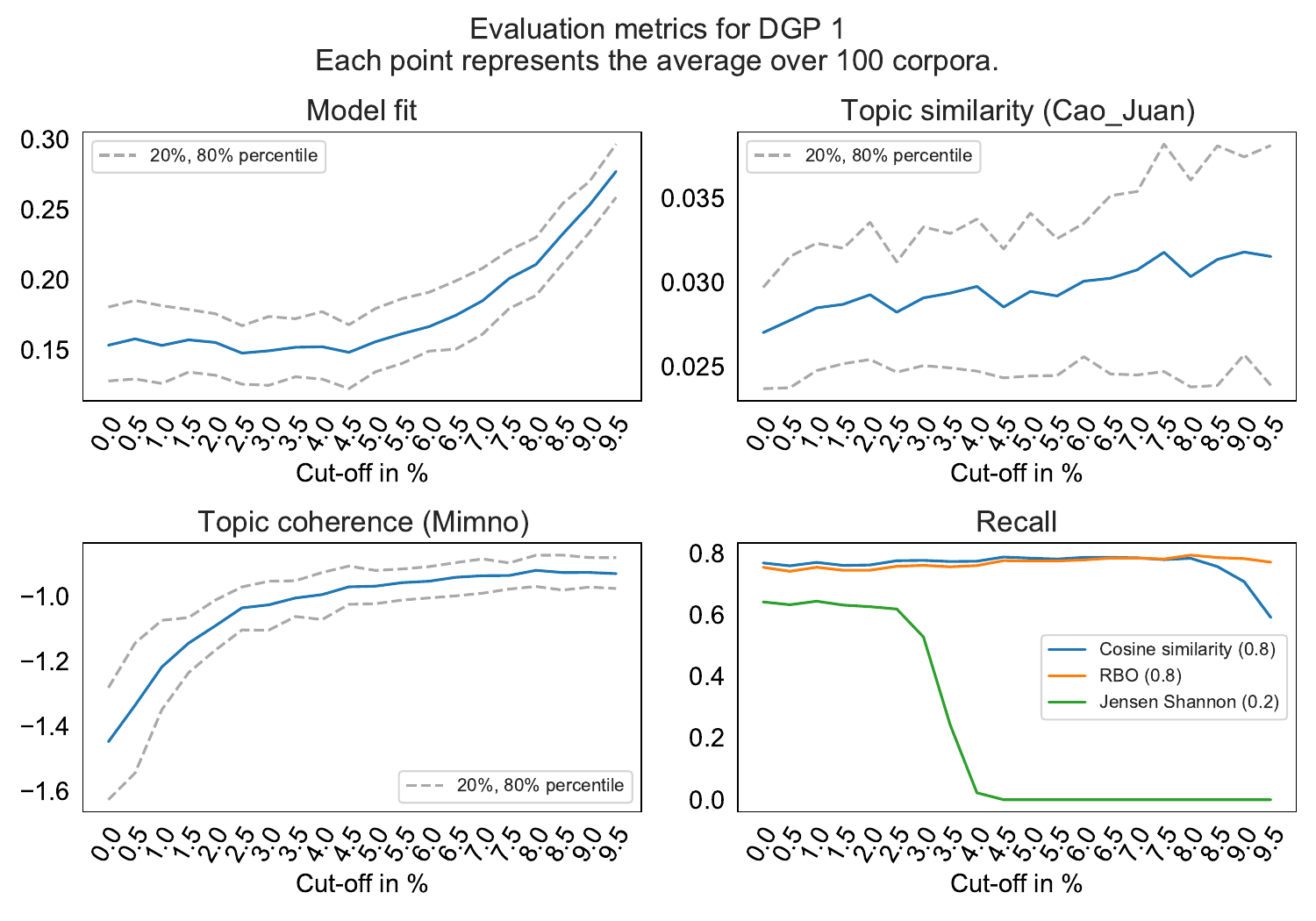"}
\caption{Evaluation of TF-IDF based vocabulary pruning for DGP1}
\label{fig:dgp1_metrics_tfidf}
\end{figure}

\begin{figure}[H]
\centering
\includegraphics[width=0.85\textwidth]{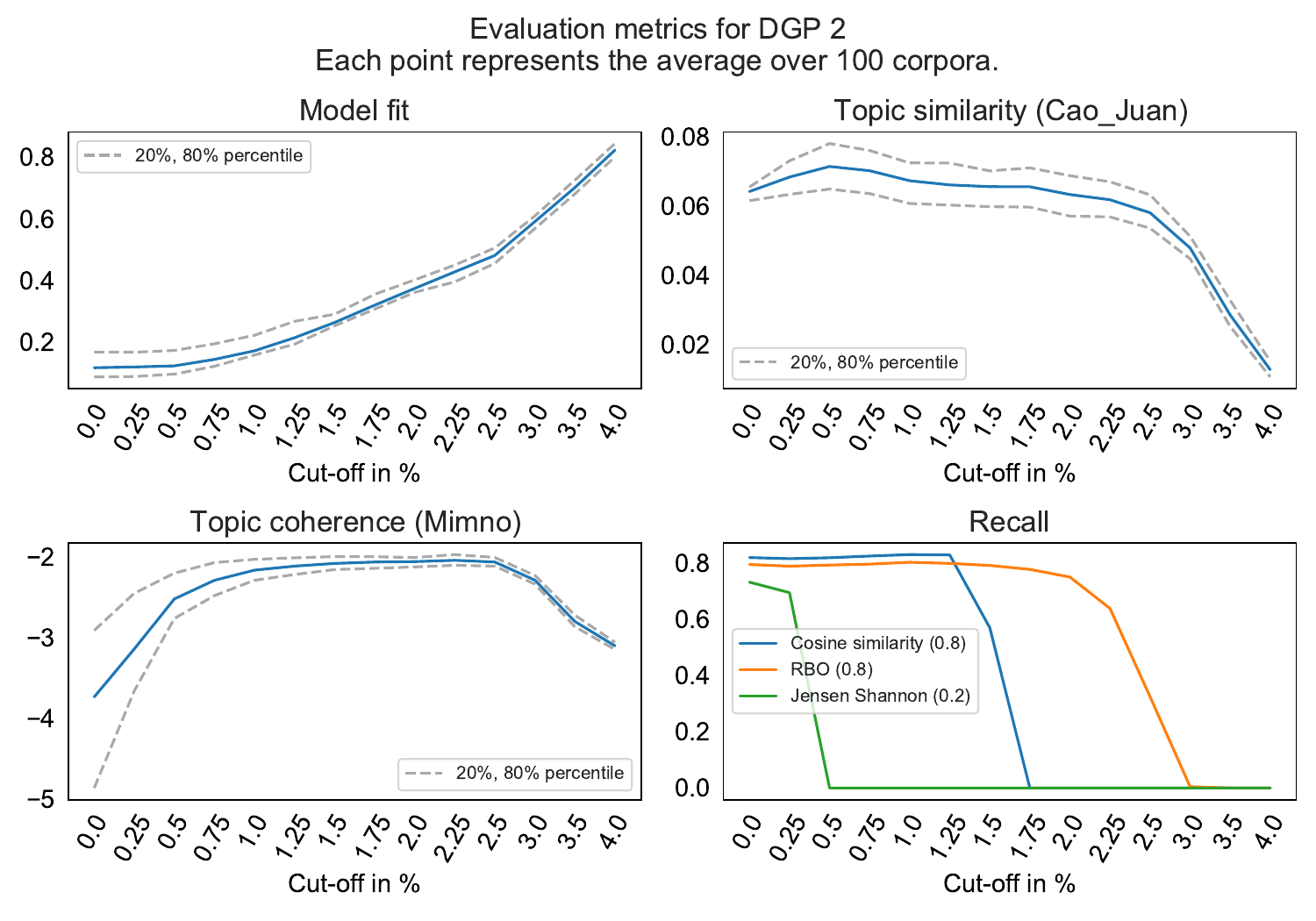}
\caption{Evaluation of TF-IDF based vocabulary pruning for DGP2}
\label{fig:dgp2_metrics_tfidf}
\end{figure}

\begin{figure}[H]
\centering
\includegraphics[width=0.85\textwidth]{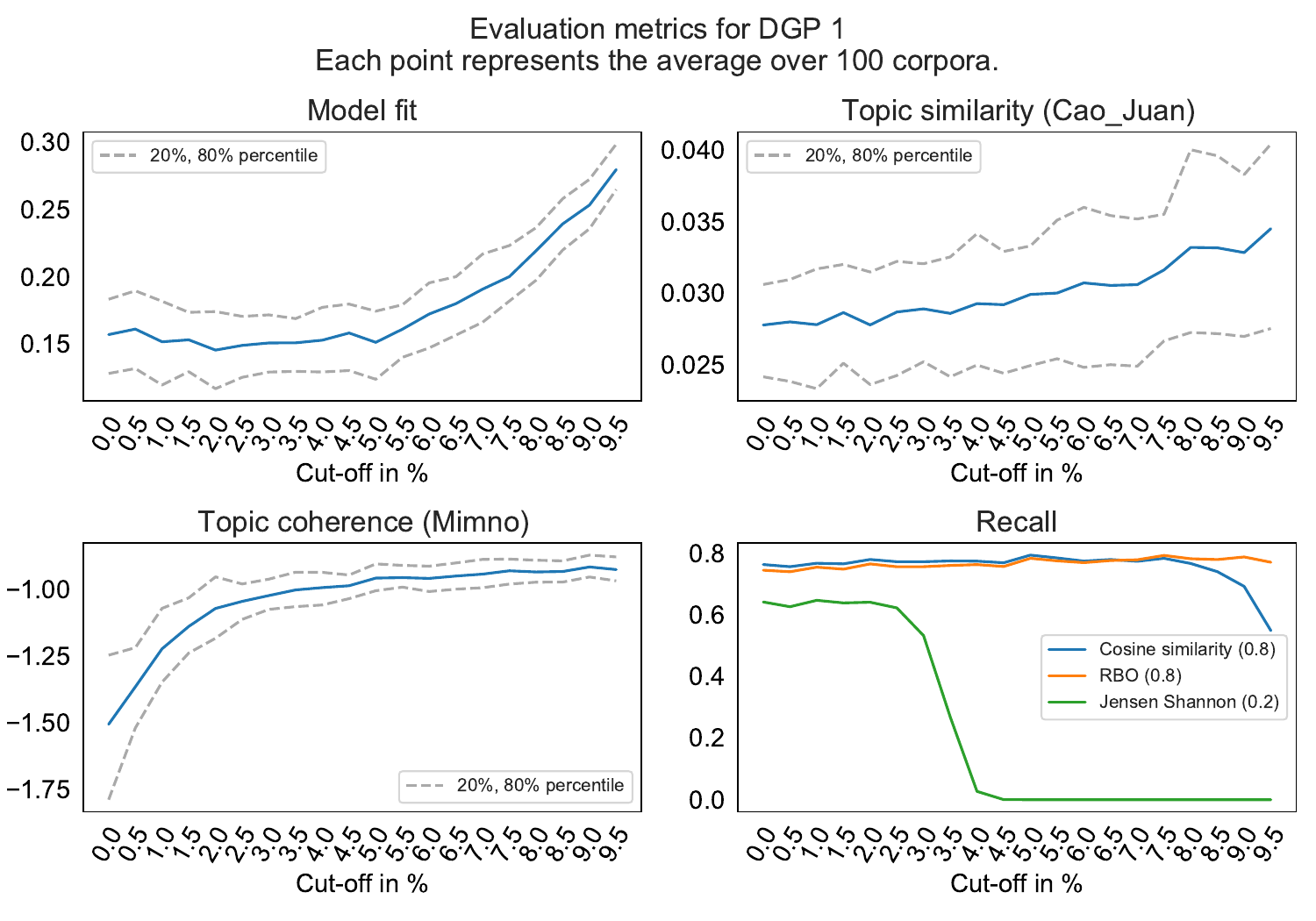}
\caption{Evaluation of absolute term frequency based vocabulary pruning for DGP1}
\label{fig:dgp1_metrics_absolute}
\end{figure}

\begin{figure}[H]
\centering
\includegraphics[width=0.85\textwidth]{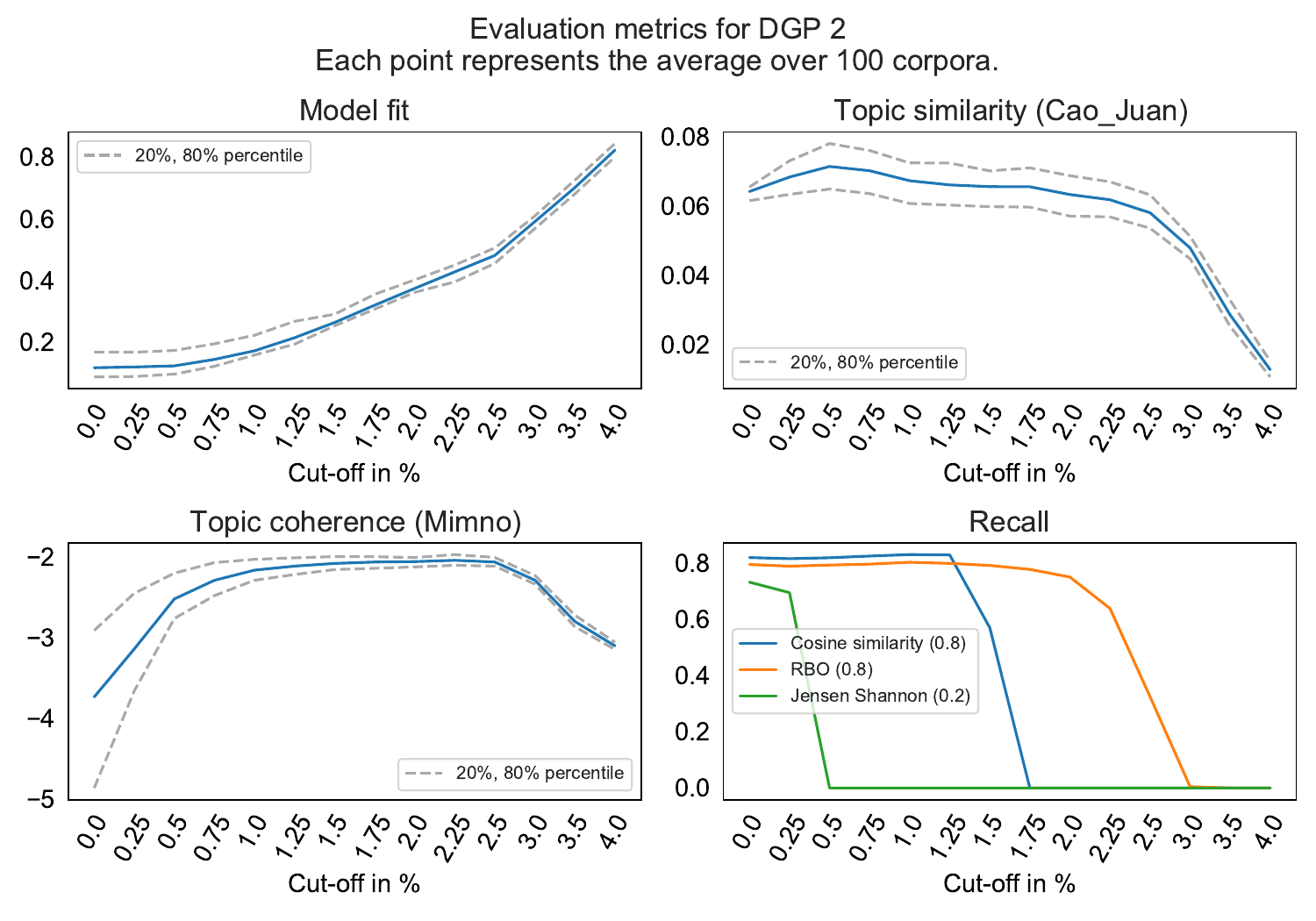}
\caption{Evaluation of absolute term frequency based vocabulary pruning for DGP2}
\label{fig:dgp2_metrics_absolute}
\end{figure}

Figures~\ref{fig:dgp1_metrics_tfidf} -- \ref{fig:dgp2_metrics_absolute} provide results when using alternative metrics for defining low-frequency terms. In Figures~\ref{fig:dgp1_metrics_tfidf} and ~\ref{fig:dgp2_metrics_tfidf}, the exclusion is based on the TF-IDF values of the terms, while the absolute frequency of terms in a corpus is used for Figures~\ref{fig:dgp1_metrics_absolute} and~\ref{fig:dgp2_metrics_absolute}.

\section{Additional Visualizations}\label{appendixb}
\begin{figure}[H]
\centering
\includegraphics[width=0.85\textwidth]{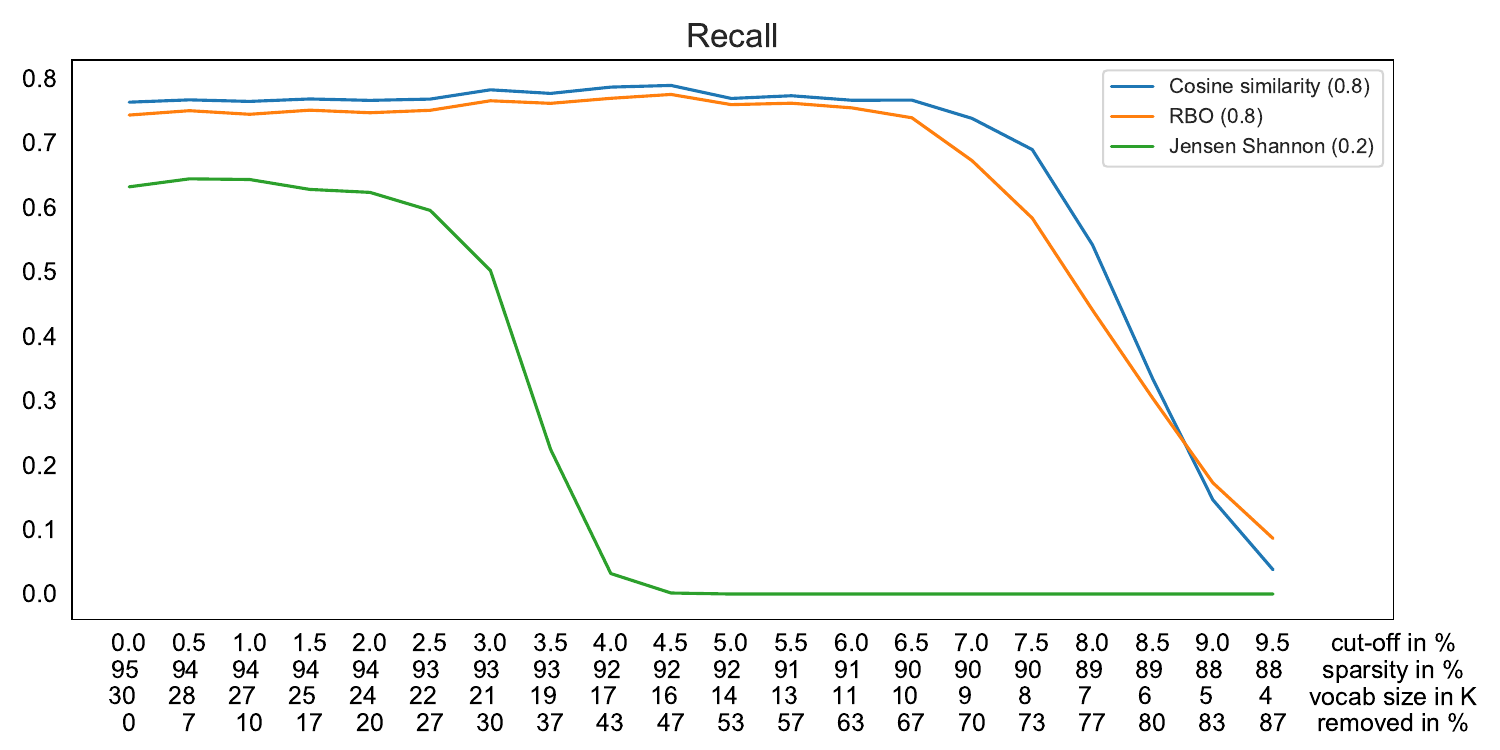}
\caption{Recall values and additional statistics for DGP1}
\label{fig:dgp1_recall_add_axes}
\end{figure}

\begin{figure}[H]
\centering
\includegraphics[width=0.85\textwidth]{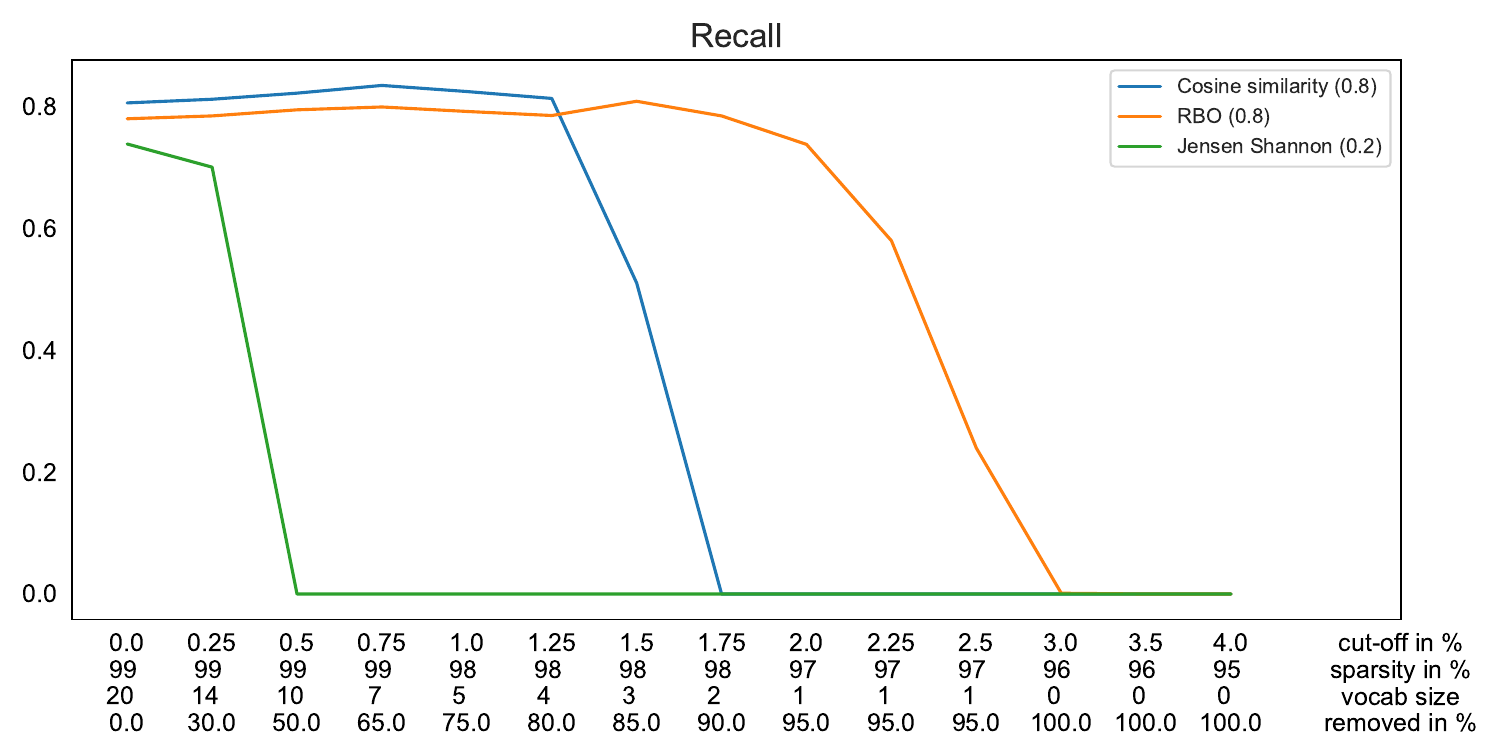}
\caption{Recall values and additional statistics for DGP2}
\label{fig:dgp2_recall_add_axes}
\end{figure}
\begin{figure}[H]
    \centering
    \includegraphics[width=1.2\textwidth]{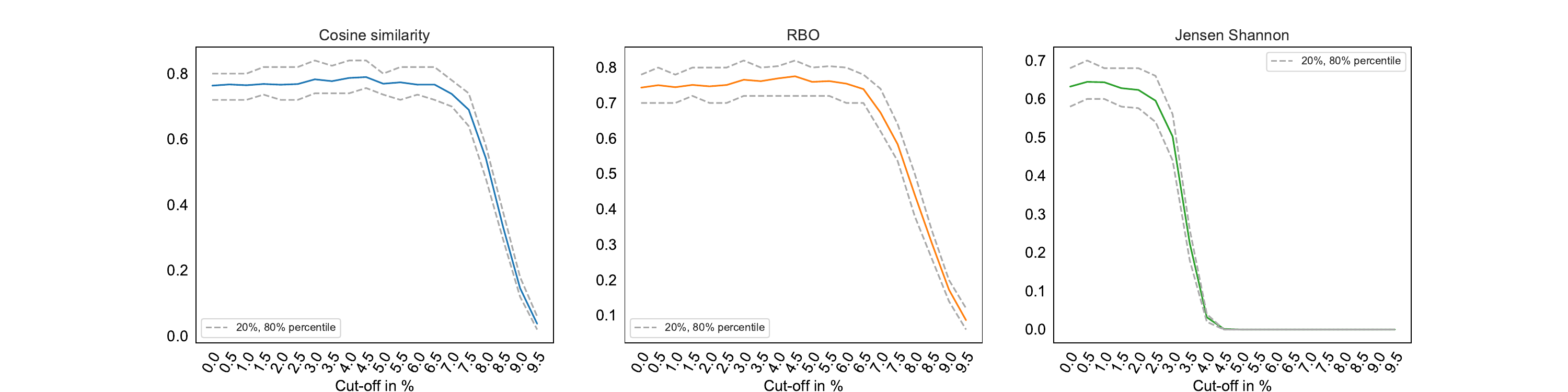}
    \caption{Recall values for DGP1}
    \label{fig:dgp1_recall_bands}
\end{figure}

\begin{figure}[H]
    \centering
    \includegraphics[width=1.2\textwidth]{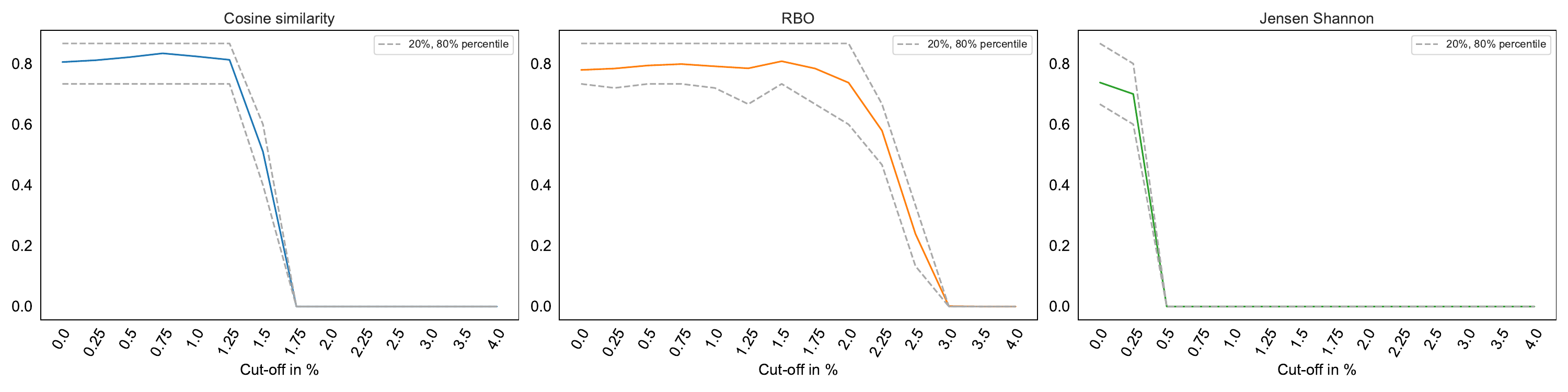}
    \caption{Recall values for DGP2}
    \label{fig:dgp2_recall_bands}
\end{figure}
\end{appendices}
\end{document}